# A Large Language Model for Feasible and Diverse Population Synthesis

Sung Yoo Lim[1†], Hyunsoo Yun[2†], Prateek Bansal[3], Dong-Kyu Kim[2,4] and Eui-Jin Kim[1*]

[1]Department of Transportation Systems Engineering, Ajou University, Republic of Korea

[2]Institute of Construction and Environmental Engineering, Seoul National University, Republic of Korea

[3]Department of Civil and Environmental Engineering, National University of Singapore, Singapore

[4]Department of Civil and Environmental Engineering, Seoul National University, Republic of Korea

[*]Corresponding Author (euijin@ajou.ac.kr)

[†]These authors contributed equally to this work.

**Abstract**

Generating a synthetic population that is both feasible and diverse is crucial for ensuring the validity of downstream activity schedule simulation in activity-based models (ABMs). While deep generative models (DGMs), such as variational autoencoders and generative adversarial networks, have been applied to this task, they often struggle to balance the inclusion of rare but plausible combinations (i.e., sampling zeros) with the exclusion of implausible ones (i.e., structural zeros). To improve feasibility while maintaining diversity, we propose a fine-tuning method for large language models (LLMs) that explicitly controls the autoregressive generation process through topological orderings derived from a Bayesian Network (BN). Experimental results show that our hybrid LLM-BN approach outperforms both traditional DGMs and proprietary LLMs (e.g., ChatGPT-4o) with few-shot learning. Specifically, our approach achieves approximately 95% feasibility—significantly higher than the ~80% observed in DGMs—while maintaining comparable diversity, making it well-suited for practical applications. Importantly, the method is based on a lightweight open-source LLM, enabling fine-tuning and inference on standard personal computing environments. This makes the approach cost-effective and scalable for large-scale applications, such as synthesizing populations in megacities, without relying on expensive infrastructure. By initiating the ABM pipeline with high-quality synthetic populations, our method improves overall simulation reliability and reduces downstream error propagation. The source code for these methods is available for research and practical application.

# 1. Introduction

Agent-based models are widely used in travel demand modeling as they allow for the explicit representation of heterogeneous agent behaviors and inter-agent dependencies—key elements for capturing complex mobility systems and dynamic travel behavior (Rezvany et al., 2023; W. Axhausen et al., 2016). A specific subclass of agent-based models, activity-based model (ABM), is popular in travel demand modeling. ABMs has two key components – generating synthetic population of the study area and generating daily activity schedules of synthetic agents. This study focuses on the first component – *population synthesis*, which involve generating synthetic residents in a city with socio-demographic characteristics such as age, income, employment status, and household composition (Castiglione et al., 2015). An ideal synthetic population replicates the joint distribution of all socio-demographic characteristics of the city's actual population.

Typically, two data sources are used for generating synthetic populations—census data and household travel survey (HTS) data. Large-scale census data provide only marginal distributions over a limited set of attributes (e.g., age, gender, income) and do not capture joint distributions across multiple attributes, which is necessary for accurately representing synthetic population. In contrast, small-scale HTS data cover only a small fraction of the population (typically 1–5%) and therefore cannot fully represent the diversity of attribute combinations found in the actual population (Kim and Bansal, 2023). In other words, the generated synthetic population tends to overlook minority groups that are often not represented in HTS samples (e.g., elderly people living alone in rural areas), which poses a barrier to equitable and responsible travel demand modeling (Pereira et al., 2017; Zheng et al., 2021). To address limitations inherent to HTS data, simulation-based approaches using generative models have been introduced to learn the underlying joint probability distribution of individual attributes from the small-scale sample data and simulate new, unobserved attribute combinations (Farooq et al., 2013; Saadi et al., 2016; Sun and Erath, 2015). Among them, a prominent class is deep generative models (DGMs), which leverage neural network architectures—such as variational autoencoders (VAEs) and generative adversarial networks (GANs) — to capture complex, high-dimensional dependencies in the data (Borysov et al., 2019; Garrido et al., 2020; Jutras-Dubé et al., 2024; Kim and Bansal, 2023)

In the simulation-based approaches, individual attribute combinations can be classified into four types based on their presence in the sample distribution, population distribution, and generated distribution (Kim and Bansal, 2023), as shown in **Figure 1**. (i) *General samples*, which are feasible and exists in the sample, population and the generated distribution; (ii) *Missing samples*, which are feasible and exists in the sample and population distributions but are missing from the generated distribution; (iii) *Sampling zeros*, which are feasible and exists in the population and generated distribution but are missing in the sample distribution (e.g., minority group not represented in HTS); (iv) *Structural zeros*, which are infeasible that do not exist in the sample and population distributions but appear in the generated distribution (e.g., a six-year-old person owning a car).

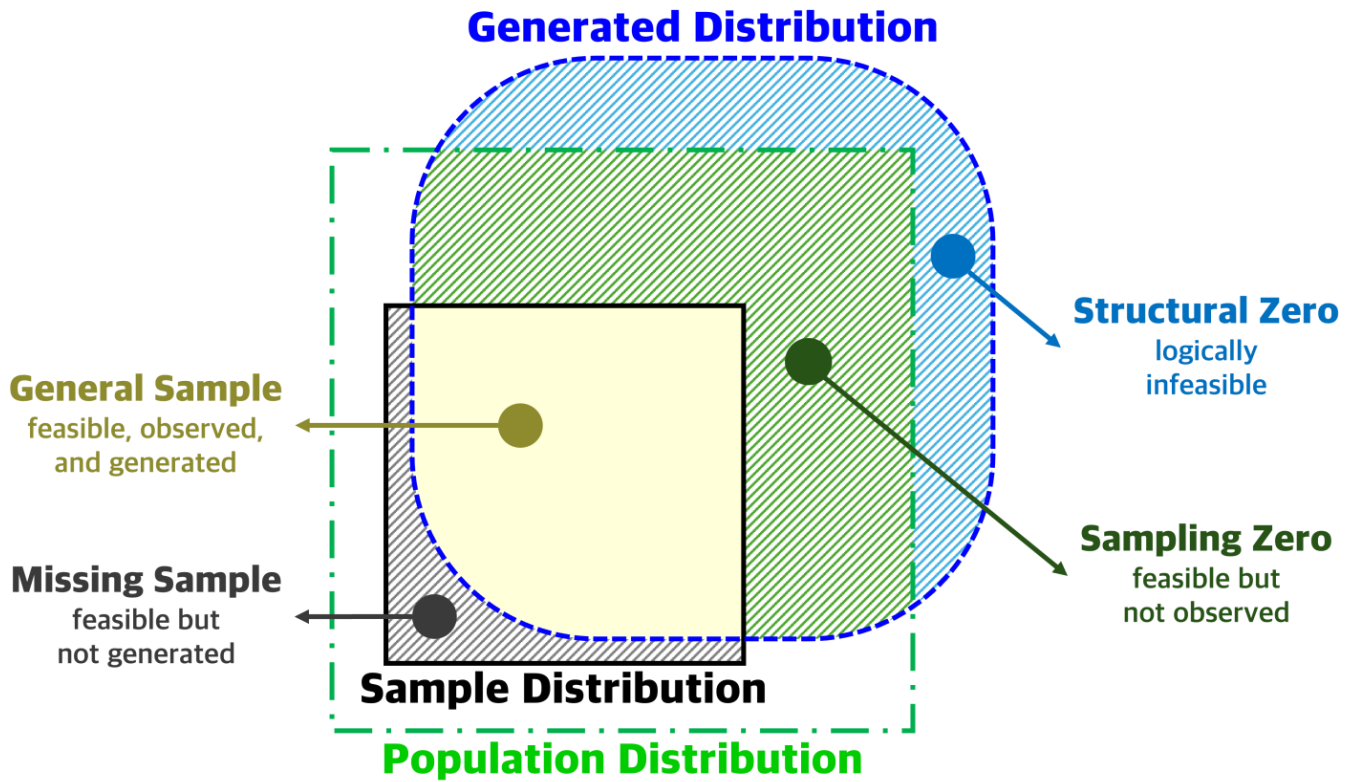

**Figure 1**. Conceptual framework of structural zeros and sampling zeros in population synthesis

It is worth noting that all generative models inherently generate structural and sampling zeros. For example, in the context of large language models (LLMs), structural zeros are closely related to the concept of *hallucination* (Huang et al., 2025)—referring to implausible outputs that do not align with real-world logic. In contrast, sampling zeros reflect a model's capacity for *creativity* (Franceschelli and Musolesi, 2024): the ability to generate plausible but previously unobserved attribute combinations that go beyond the training data.

An ideal synthetic population aims to maximize the inclusion of sampling zeros and general samples while minimizing the missing samples and structural zeros, thereby closely approximating the true population distribution. However, this trade-off remains a fundamental challenge—pushing the generative model to produce new attribute combinations beyond those observed in the training data (i.e., sampling zeros) increases the likelihood of generating implausible ones (i.e., structural zeros). This trade-off can be alleviated, i.e., a higher number of sampling zeros can be generated without generating many structural zeros, by considering the semantic meaning and associations between attributes. Such semantic relationships can help identify the attribute combinations that are typical, rare, or infeasible. However, such considerations are difficult to implement in statistical or deep learning approaches used for population synthesis (e.g., DGMs), especially when using small-scale samples.

To address these challenges, we propose a novel approach for synthetic population generation that leverages LLMs as they have demonstrated a strong ability to capture semantic relationships between attributes (Wei et al., 2022). These relationships are inferred from statistical patterns embedded in diverse and large-scale text corpora, setting LLMs apart from conventional generative models that primarily rely on observed joint distributions from limited training data. We hypothesize that LLMs can encode semantic relationships between socio-demographic attributes even in regions of the attribute space that are sparsely represented or entirely absent from the training data. By fine-tuning LLMs on the small-scale training data, domain-specific knowledge embedded in those datasets can be reinforced and aligned with the broader semantic relationships acquired during pretraining. This property makes LLMs particularly well-suited for minimizing the structural zeros while preserving the sampling zeros—without the need for manually defined constraints or infeasibility rules.

Several recent studies have integrated LLMs into travel demand models, such as activity scheduling decisions involving location, departure time, and travel mode (Liu et al., 2024a, 2024b; Wang et al., 2024). While their findings highlight the potential of LLMs to model human behavior in ABMs, they largely overlook the task of generating feasible and diverse synthetic populations.

In contrast to prior studies that use LLM for modeling travel behavior, our approach leverages LLMs as semantically informed population generators. Our initial empirical findings suggest that population synthesis remains a non-trivial task for off-the-shelf proprietary LLMs (e.g., GPT-4o), as their sequential generation results in reduced diversity and feasibility of the synthesized population. Therefore, we fine-tune a pretrained lightweight open-source LLM by guiding it toward semantically plausible outputs using a Bayesian Network (BN) that captures conditional dependencies among attributes in a data-driven manner. We integrate the BN as it generally performs on par with state-of-the-art DGMs in generating diverse and feasible synthetic populations (Kim and Bansal, 2023), while offering superior interpretability that makes it well-suited for guiding LLMs. Specifically, during fine-tuning, the BN provides a topological ordering that structures the input text and reflects realistic attribute hierarchies. During generation, the BN-derived ordering is used to structure the initial prompt, guiding the LLM to generate attributes in a semantically coherent manner.

Our contributions are threefold. First, we quantitatively evaluate the performance of proprietary LLMs (e.g., GPT-4o) with few-shot learning (Liu et al., 2024a, 2024b) on the population synthesis task in terms of feasibility and diversity (Kim and Bansal, 2023). This baseline demonstrates the limitation of LLMs when applied to population synthesis without fine-tuning.

Second, we propose a novel fine-tuning method for synthesizing a population that explicitly controls the autoregressive generation process by shaping the attribute search space during training, balancing the trade-off between feasibility and diversity. This is achieved by incorporating topological orders—i.e., conditional dependencies among attributes—learned from a BN, and controlling the decoding temperature and fine-tuning depth through the number of training epochs.

Third, we show that the proposed method significantly improves the overall quality of the synthetic population, particularly in terms of feasibility, compared to existing DGMs. Notably, these improvements are achieved by fine-tuning a lightweight open-source LLM and generating synthetic population on standard personal computers. This eliminates dependency on proprietary or large-scale LLM, which typically incur high computational or API-related token costs. These advantages of the proposed approach become even more evident for ABMs, as they typically require large-scale population synthesis. For example, in travel demand modeling, it is often necessary to generate synthetic populations for large metropolitan areas—such as 10 million individuals in a city like Seoul.

The remainder of this paper is organized as follows. Section 2 reviews the relevant literature on population synthesis, including traditional methods, deep generative models, and recent efforts to leverage large language models for generating structured data. Section 3 introduces the dataset used in this study. Section 4 outlines the proposed methodology, which includes the structure-informed fine-tuning strategy and the generation process. Section 5 presents the experimental results and evaluates the effectiveness of the proposed approach. Finally, Section 6 concludes the paper and discusses directions for future research.

## 2. Related Work

Population synthesis is a multi-stage process involving (i) generation of synthetic individuals with socio-demographic characteristics , and (ii) spatial allocation of home and work locations of synthetic individuals in the study area (Bigi et al., 2024; Borysov et al., 2019; Vo et al., 2025). This study focuses on the first stage—generating diverse yet feasible synthetic individuals from a limited sample (e.g., HTS). For a comprehensive review of the population synthesis framework, readers are referred to La et al. (2025).

### 2.1 Generative Modeling Approaches for Population Synthesis

A central challenge in synthesizing population is to construct individual-level agents whose joint distribution of attributes resemble those of the true population. This is challenging to achieve using small-scale HTS samples with missing attribute combinations, i.e., zero probability mass on several attribute combinations (i.e., zero cell problem). Generative modeling approaches address this by learning the underlying data distribution and sampling from it to produce plausible attribute combinations (i.e., sampling zeros) that are unobserved in HTS data. Early methods often relied on statistical modeling techniques such as BN (Sun and Erath, 2015), Markov chain Monte Carlo (Farooq et al., 2013), or Hidden Markov Model (Saadi et al., 2016), which offer explicit control over attribute dependencies but often require handcrafted structures or strong independence assumptions. These approaches struggle with scalability issues when synthesizing high-dimensional attribute spaces or recovering rare attribute combinations not observed in the training data (Borysov et al., 2019).

To improve the richness of synthetic population with the diverse combinations of socio-demographic attributes, recent studies have adopted DGMs, such as VAEs and GANs (Badu-Marfo et al., 2022; Borysov et al., 2019; Garrido et al., 2020; Kim and Bansal, 2023). DGMs are particularly effective in addressing the zero-cell problem (Bigi et al., 2024; Choupani and Mamdoohi, 2016; Guo and Bhat, 2007). By learning complex interdependencies among attributes, DGMs can extrapolate beyond the observed data space and generate such sampling zeros, thus enhancing the diversity of synthetic populations.

Within the broader machine learning community, several DGM variants have been proposed for tabular data generation, a task closely related to population synthesis. For example, DATGAN (Lederrey et al., 2022) incorporates expert knowledge through a DAG to guide training; CTAB-GAN (Zhao et al., 2021) focuses on effectively modeling both discrete and continuous attributes while addressing data imbalance; and TabDDPM (Kotelnikov et al., 2023) demonstrates the potential of diffusion models for generating tabular dataset.

Although DGMs can generate attribute combinations that are missing in the sample but exist in the population (i.e., sampling zeros), they often do so at the cost of generating infeasible combinations that do not exist in the true population (i.e., structural zeros). Most DGMs lack an explicit mechanism to distinguish between sampling and structural zeros. Kim and Bansal (2023) proposed regularization-based modifications to both VAEs and GANs to explicitly control the trade-off between sampling and structural zeros.

This study builds upon literature on generative models and achieves substantial improvements in synthesizing population by reducing structural zeros while preserving the sampling zeros through the incorporation of LLMs.

## 2.2 Large Language Models for Travel Demand Modeling

The applications of LLMs are growing in travel demand modeling. Early efforts have focused on leveraging the generative and reasoning capabilities of LLMs to emulate humans' travel behavior, particularly within ABM frameworks (Liu et al., 2024b, 2024a; Qin et al., 2025). Notably, Liu et al. (2024a) propose a conceptual framework where LLMs are embedded as agents capable of capturing complex decision-making process underlying human travel behavior. Similarly, Wang et al. (2024) introduce LLMob, a personal mobility generation system that models daily activity patterns using structured prompts, templates, and motivation retrieval mechanism.

While these studies demonstrate the potential of using LLMs to generate agent behavior within ABMs, they overlook the feasibility and diversity of the synthetic agents (i.e., the synthetic population). Since individual attributes of synthetic population (e.g., sociodemographic attributes) influence downstream modules of ABM, even minor inconsistencies in these attributes can propagate through the simulation process, leading to substantial errors in outcome. Zhang et al., (2024) observed that LLMs, when used in prompt-only configurations, often fail to produce realistic activity patterns without fine-tuning, highlighting the fragility of behavioral outputs when LLMs are not sufficiently aligned with domain-specific constraints. These findings highlight the importance of carefully managing the integration of LLMs into the ABM pipeline.

In this regard, our work introduces LLMs at an earlier stage of the ABM—generating synthetic population. We leverage semantic priors of LLMs to generate a feasible and diverse synthetic population. Specifically, by minimizing the structural zeros while preserving sampling zeros in synthesizing population, our approach contributes to a more robust and controllable foundation for ABM simulations.

Prior work related to the proposed method is GReaT (Borisov et al., 2023), which uses LLMs to generate individual-level tabular data, aligning with the goal of population synthesis. While GReaT shows strong performance across benchmark datasets in the original paper, its application in population synthesis for ABM remains unexplored. GReaT introduces a fine-tuning approach for LLMs on text-formatted tabular rows with randomly permuted feature orders. This exposes the model to randomized autoregressive structures, forcing it to cover the entire attribute space during training. However, in contexts such as ABMs, where attributes exhibit semantic relationships (e.g., age → employment status → income), such randomized ordering create an excessively large attribute space, leading to inefficient training. Building on this insight, this study focuses on developing a fine-tuning method that enables explicit control over the autoregressive generation process in LLMs, with the goal of balancing feasibility and diversity in synthetic population generation. We frame this control around two stages: (i) the efficient training, in which the attribute search space is constrained using a topological attribute ordering derived from a BN to structure input sequences and regulate the number of fine-tuning epochs; and (ii) the controllable generation, where decoding temperature influences the randomness of model's output. These components together manage the trade-off between diversity and feasibility in LLM-based population synthesis.

## 3. Dataset

Observing extensive individual-level attribute combinations is typically infeasible in many countries due to limited sample sizes, which often range from 30,000 to 150,000 individuals (Sun and Erath, 2015; Borysov et al., 2019; Bigi et al., 2024). In contrast, we use a dataset constructed by merging the South Korean HTS conducted in 2010, 2016, and 2021, resulting in over one million records. Given this substantially larger sample size compared to most datasets used in the literature, we assume that the presence of sampling zeros in our data is minimal. The dataset comprises 13 socio-demographic and mobility-related attributes, each of which is discretized into categorical variables, resulting in a total of 70 categories. This discretization enables us to represent the population distribution as a finite set of attribute combinations, allowing for the exact computation of evaluation metrics such as precision and recall, which we further explain in the methodology section (Section 4.4).

Since the true population is unknown, we adopt a validation strategy in synthetic population literature (Kim and Bansal, 2023; Vo et al., 2025), leveraging our large-scale dataset. Specifically, the full dataset is treated as a hypothetical population (h-population), and a 5% random subset (approximately 50,000 records) is designated as a hypothetical sample (h-sample). During training, only the h-sample is used, mimicking real-world sample size limitations of HTS data (Habib et al., 2020). **Figure 2** shows the number of attribute combinations missing in the h-sample relative to the h-population across different sampling rates. In a 5% random sample, only 11.7% of the attribute combinations present in the h-population are observed in the h-sample, covering 56.4% of the total data instances. This result highlights the severity of the zero-cell problem in traditional HTS datasets (h-sample in this case). The h-population serves as a proxy ground truth for evaluating the model's generalization ability, particularly in terms of its capacity to recover rare but plausible attribute combinations (sampling zeros) and suppress implausible attribute combinations (structural zeros). The descriptive statistics of the h-population are shown in **Table 1**.

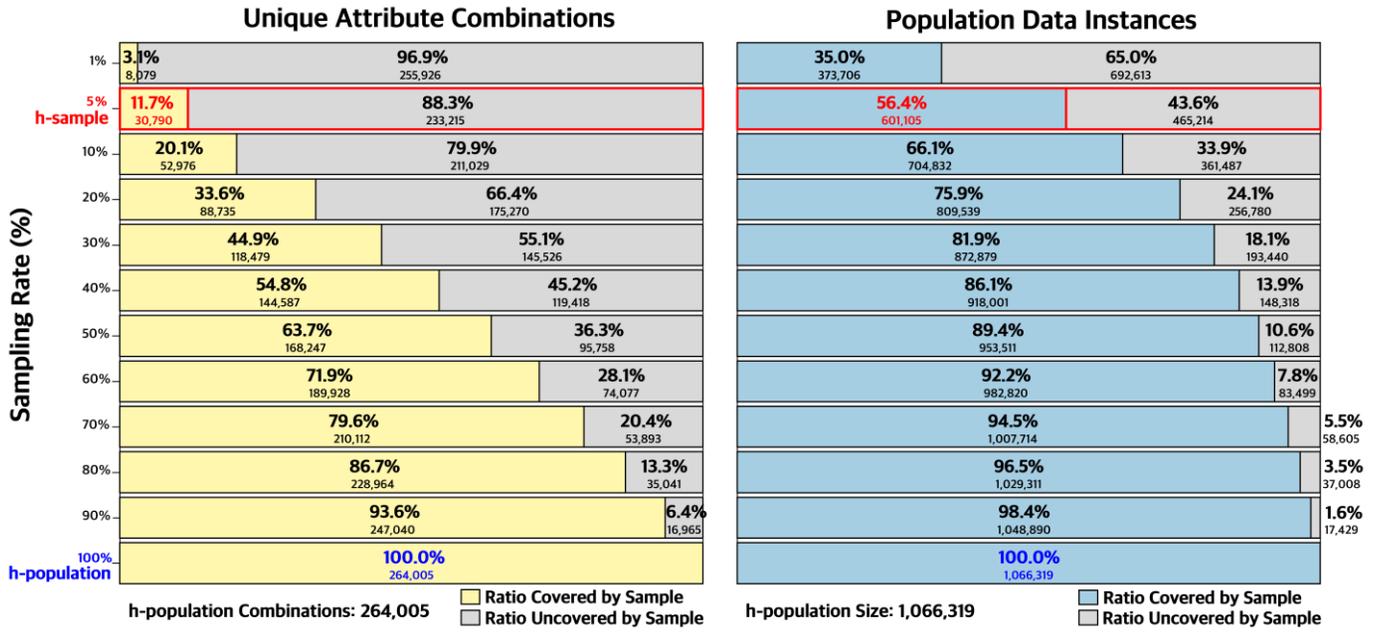

**Figure 2**. Coverage of unique attribute combinations and data instances by sampling rates

Under the assumption that the h-population is a representative approximation of the true population, the relative diversity between the h-sample and h-population can serve as a meaningful analog to the diversity gap between an HTS sample and the unknown real-world population. Consequently, the evaluation results derived from this semi-synthetic setting can be interpreted as a generalizable measure of the model performance (Kim and Bansal, 2023).

**Table 1** Description of the h-population (N =1,006,391)

| Attribute (Dimensions) | Category | Proportion (%) | Category | Proportion (%) |
|---|---|---|---|---|
| Household Monthly Income Level (5) | < 1 million KRW | 8.47 | 5–10 million KRW | 16.09 |
|  | 1–3 million KRW | 39.46 | > 10 million KRW | 2.19 |
|  | 3–5 million KRW | 33.78 |  |  |
| Car Ownership of Household (2) | Yes | 83.91 | No | 16.19 |
| Driver License (2) | Yes | 60.13 | No | 39.87 |
| Gender (2) | Male | 51.23 | Female | 48.77 |
| Home Type (6) | Apartment | 55.41 | Single-family | 21.32 |
|  | Villa | 12.09 | Multi-family | 9.48 |
|  | Studio-type/residence | 0.82 | Other | 0.89 |
| Age Group (17) | 5–10 years | 4.96 | 51–55 years | 8.89 |
|  | 11–15 years | 7.59 | 56–60 years | 7.38 |
|  | 16–20 years | 7.48 | 61–65 years | 5.57 |
|  | 21–25 years | 4.96 | 66–70 years | 4.27 |
|  | 26–30 years | 6.08 | 71–75 years | 3.02 |
|  | 31–35 years | 7.03 | 76–80 years | 2.23 |
|  | 36–40 years | 9.42 | 81–85 years | 1.16 |
|  | 41–45 years | 9.90 | 86–90 years | 0.42 |
|  | 46–50 years | 9.67 |  |  |
| Work Days (4) | 5 days/week | 27.81 | 1–4 days/week | 10.05 |
|  | 6 days/week | 17.33 | Inoccupation/non-regular | 44.82 |
| Work Type (9) | Student | 15.45 | Service | 15.69 |
|  | Inoccupation/Housewife | 18.40 | Simple labor | 12.31 |
|  | Manager/Office | 11.54 | Experts | 11.07 |
|  | Sales | 5.44 | Agriculture/Fisher | 5.68 |
|  | Others | 4.43 |  |  |
| Kid in Household (2) | Yes | 11.04 | No | 88.96 |
| Number of Household Members (7) | 1 | 7.56 | 5 | 9.67 |
|  | 2 | 18.16 | 6 | 1.32 |
|  | 3 | 25.27 | 7 | 0.14 |
|  | 4 | 37.88 |  |  |
| Major Travel Mode (6) | Car | 25.65 | Taxi | 0.31 |
|  | Public transportation | 22.49 | Walking | 21.53 |
|  | Bike/Bicycle | 2.14 | None | 27.87 |
| Major Departure Time (4) | Peak | 56.38 | Non-peak | 13.45 |
|  | Others | 2.31 | None | 27.87 |
| Education Status (4) | Not student | 76.67 | Elementary/Middle/High | 18.01 |
|  | University | 4.75 | Preschool | 0.58 |

## 4. Methodology

This section presents the proposed LLM-based population synthesis model. Our method aims to fine-tune a LLM so that it can generate feasible and diverse individuals whose attribute combinations reflect both the statistical distributions in the h-sample (i.e., training data for fine-tuning) and the semantic relationships learned by a pretrained LLM. To achieve this, we first encode individual-level attributes into structured textual sequences suitable for processing by a pretrained autoregressive LLM. We then learn BN from the training data to identify conditional dependencies among attributes. The BN provides a topological ordering of features, which guides both the fine-tuning and generation processes, helping preserve conditional dependencies and reduce the attribute search space during autoregressive generation. The overall framework is illustrated in **Figure 3**. Each component of the framework is described in detail in the following subsections.

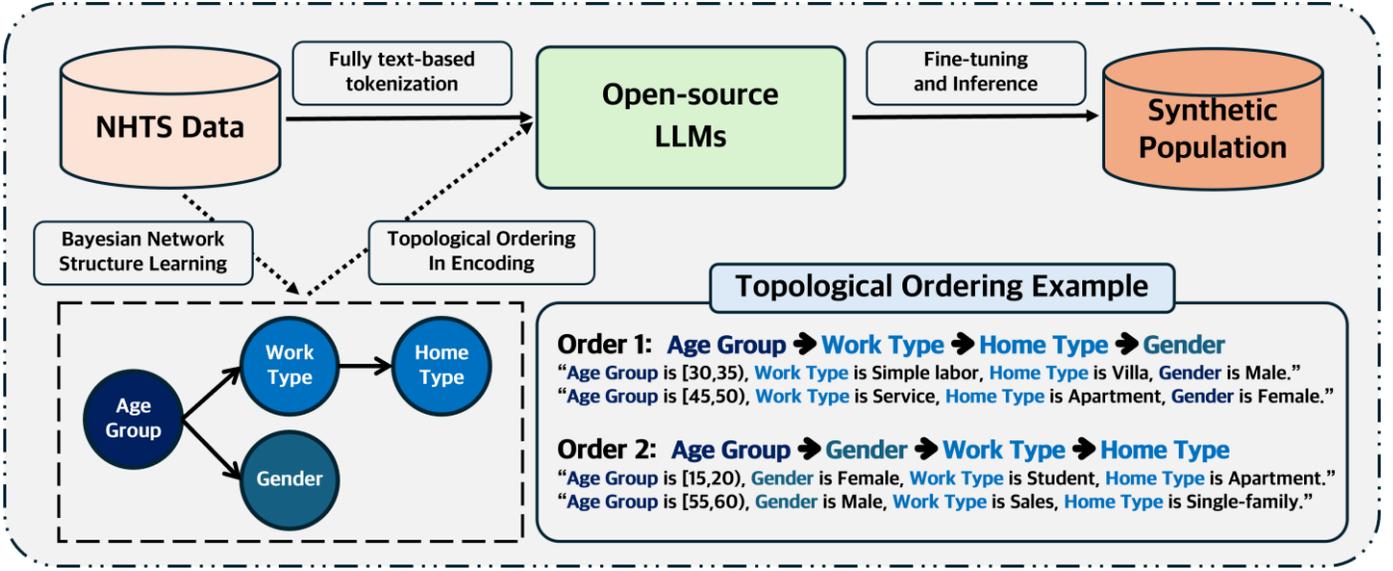

**Figure 3.** LLM Fine-tuning pipeline guided by BN-derived topological ordering.

**4.1 Fine-Tuning of Large Language Models for Population Synthesis**

Let $\mathcal{X} = \{X_1, X_2, \ldots, X_d\}$ be a set of discrete individual-level variables (e.g., age group, employment type, travel mode). Let $\mathcal{D} = \{\mathbf{x}^{(1)}, \ldots, \mathbf{x}^{(N)}\}$ be the observed dataset, where each individual is represented as a $d$-dimensional vector $\mathbf{x}^{(n)} = (x_1^{(n)}, \ldots, x_d^{(n)}) \in \mathcal{X}$. Our objective is to fine-tune (update the weights of a pretrained model on a task-specific dataset) a LLM, denoted by $P_\theta$, such that it can generate new synthetic individuals $\hat{\mathbf{x}} \sim P_\theta$ that approximate the true joint distribution $P(X_1, X_2, \ldots, X_d)$ of the population. The pre-trained LLM used in this study is a distilled version of GPT-2 (Sanh et al., 2019). This model contains approximately 82 million parameters—significantly fewer than proprietary LLMs such as GPT-3.5, which has 6.7 billion parameters (Ray, 2023)—making it feasible to fine-tune and perform large-scale generation (e.g., synthesizing one million individuals) on a standard personal computer. Let $T^{(n)} = (w_1^{(n)}, \ldots, w_{L_n}^{(n)})$ denote the textual (tokenized) sequence corresponding to the $n$-th individual. The LLM is trained to maximize the probability of each token given its preceding context, which is equivalent to minimizing the negative log-likelihood:

$$\mathcal{L}(\theta) = -\sum_{n=1}^{N} \sum_{t=1}^{L_n} \log P_\theta(w_t^{(n)} \mid w_1^{(n)}, \ldots, w_{t-1}^{(n)}) \tag{1}$$

As shown in **Equation 2**, this corresponds to a standard autoregressive factorization:

$$P_\theta(w_1, w_2, \ldots, w_L) = \prod_{t=1}^{L} P_\theta(w_t \mid w_1, \ldots, w_{t-1}) \tag{2}$$

Standard LLMs are pretrained on natural language corpora and require tokenized text as input. Therefore, to apply LLM for population synthesis, each individual record from the tabular dataset is transformed into a sequence of descriptive text units. Given an ordering of attributes $\pi = [X_{\pi(1)}, X_{\pi(2)}, \ldots, X_{\pi(d)}]$, we define a deterministic phrase template for each variable as: $e_{\pi(i)}(x_{\pi(i)}^{(n)}) =$ "The respondent's $X_{\pi(i)}$ is $\omega_{\pi(i)}(x_{\pi(i)}^{(n)})$", where $\omega_{\pi(i)}$ maps the categorical variable $x_{\pi(i)}$ to a human-readable phrase (e.g., "3" → "between 3 and 5 thousand dollars"). The full sequence $T^{(n)}$ describing individual $n$ is then constructed as:

$$T^{(n)} = e_{\pi(1)}(x_{\pi(1)}^{(n)}) + \cdots + e_{\pi(d)}(x_{\pi(d)}^{(n)}) \tag{3}$$

This transformation enables structured tabular data to be represented in a form compatible with LLMs while preserving semantic attribute meanings. The topological ordering $\pi$, determines the dependency structure learned by LLM. By explicitly controlling this ordering, we can influence how the model balances feasibility and diversity in its outputs, a central aspect of our approach.

In training conventional DGMs, a sufficient number of epochs is typically required to ensure that the model adequately learns the distribution of the training data. However, in our experiments with LLM fine-tuning, we observed that increasing the number of epochs led to greater reliance on the h-sample, which in turn reduced the model's ability to generate sampling zeros based on the semantic relationships encoded in the pretrained LLM.

To address this, we treat the number of epochs as a key hyperparameter to control the *fine-tuning depth*. As the fine-tuning depth increases, the model becomes increasingly aligned with the sample distribution, resulting in improved feasibility. However, this comes at the cost of reduced diversity, as the influence of the pretrained LLM's semantic knowledge diminishes. These findings suggest that fine-tuning depth serves as a useful hyperparameter for managing the trade-off between feasibility and diversity in LLM-based population synthesis.

## 4.2 Structure Learning with Bayesian Networks

In actual populations, many attributes exhibit strong conditional dependencies. For example, a child is unlikely to be employed, and high-income individuals are more likely to own a car. Therefore, the choice of ordering $\pi$ significantly affects the autoregressive model's ability to learn dependencies among attributes. To encode the dependencies among attributes and obtain a principled ordering, we utilize the BN. A BN is defined as $\mathcal{B} = (\mathcal{G}, \phi)$ where $\mathcal{G} = (\mathcal{V}, \mathcal{E})$ is a directed acyclic graph (DAG) over a set of nodes corresponding to attributes $\mathcal{V} = \{X_1, \ldots, X_d\}$, $\mathcal{E}$ is the set of directed edges between these nodes, and $\phi$ is the set of conditional probability tables associated with each node. The BN factorizes the joint distribution as:

$$P(X_1, X_2, \ldots, X_d) = \prod_{i=1}^{d} P(X_i \mid \text{Pa}(X_i)) \tag{4}$$

where $\text{Pa}(X_i)$ denotes the set of parent variable of $X_i$ in the DAG $\mathcal{G}$. To estimate the structure $\mathcal{G}$ (i.e., conditional dependencies among attributes), we apply score-based structure learning using Hill-Climb Search (Tsamardinos et al., 2006). Given the discrete nature of the data, we adopt the K2 score metric (Cooper and Herskovits, 1992) to guide the search process.

To align with the left-to-right autoregressive structure of LLMs, we restrict the BN to have a maximum in-degree 1, i.e., each node has at most one parent. This induces a chain-like DAG, enabling a topological ordering such that:

$$X_j \in \text{Pa}(X_i) \Rightarrow j < i \text{ in } \pi \tag{6}$$

This constraint ensures that the BN admits a valid topological ordering $\pi = [X_{\pi(1)}, \ldots, X_{\pi(d)}]$ that aligns with the sequential nature of autoregressive generation. During fine-tuning, we dynamically sample valid topological orders for each training instance to enhance robustness and generalization across different attribute permutations. Notably, when the learned DAG has multiple disconnected chains, we process each chain fully before moving to the next zero in-degree node (i.e., a node with no incoming edges). This ensures acyclicity and guarantees that each attribute is conditioned only on its structural predecessors. This structure allows the model to generate attributes (e.g., household income) conditioned on structurally relevant variables (e.g., educational status, age), thereby reducing the likelihood of implausible combinations and improving behavioral realism in the generated synthetic population.

**Figure 4** illustrates an example of how the topological ordering of attributes is constructed from the learned DAG. The ordering process begins with a traversal from a root node (i.e., a node without parents)—in this example, education status—and follows one child node at a time to build a path, respecting the conditional dependencies encoded in the DAG. When a path terminates (i.e., reaches a leaf node or no further child is selected), a new traversal begins from another unvisited root node. This recursive traversal continues until all connected components have been explored. Finally, any remaining unvisited nodes—typically those not encountered during traversal—are appended in random order (e.g., Driver's License, Work Days, and Gender in this example). The resulting topological order is then used to structure the input sequence during fine-tuning, ensuring that each attribute is conditioned on its most relevant predecessors.

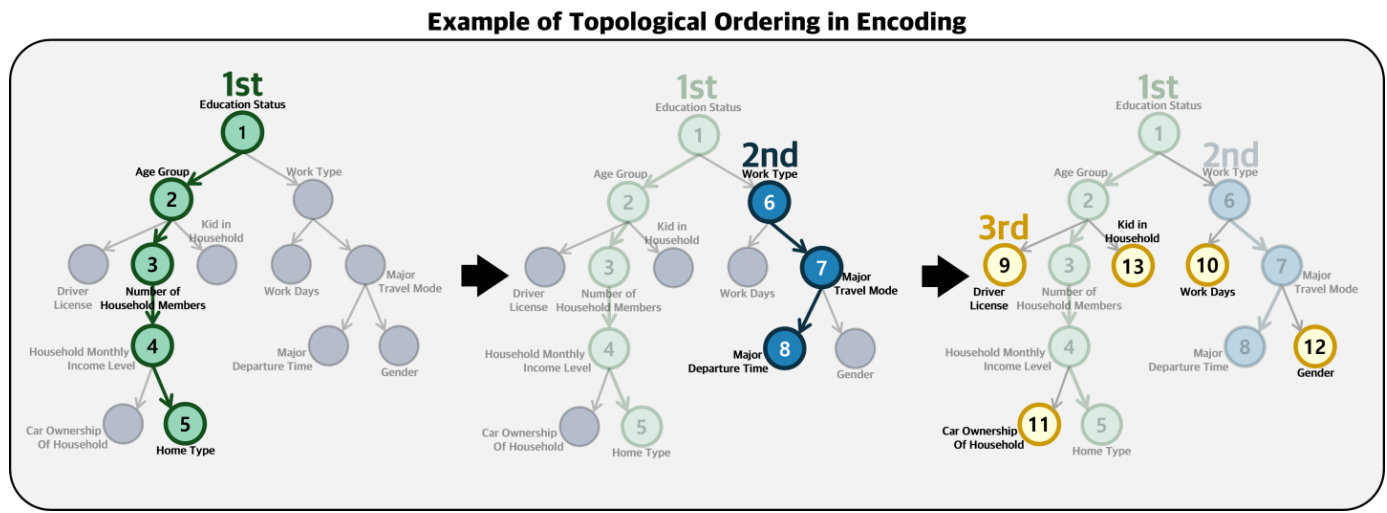

**Figure 4.** Example of topological ordering in encoding for fine-tuning.

## 4.3 Sampling from the fine-tuned LLM to generate synthetic pools

After fine-tuning, we generate synthetic individuals by sampling from the model in a sequential, autoregressive manner. No fixed topology is imposed during generation. Instead, the generation is initiated by providing the textual representation of the first attribute of DAG as a starting prompt, while the LLM completes the sequence, based on its learned distribution. Let the generation prompt be "The respondent's $X_i$ is". Subsequent tokens $w_t$ are sampled autoregressively: $w_t \sim P_\theta(w_t \mid w_1, \ldots, w_{t-1})$ until the end of the sentence. The final sequence $T = (w_1, \ldots, w_L)$ is then parsed back into a structured tabular format, yielding a synthetic population $\hat{x} \in \mathcal{X}^d$. To control the diversity of generated output, general auto-regressive LLMs implement temperature-controlled decoding, defined as:

$$P_\theta(w_t = v_i \mid w_{<t}) = \frac{exp\ (z_{t,i}/\tau)}{\sum_{j=1}^{|S|} exp\ (z_{t,j}/\tau)} \tag{7}$$

where $z_{t,i}$ is the raw logit for token $v_i$ at position $t$, $|S|$ is the size of the model's vocabulary (i.e., the number of distinct tokens the LLM can generate), and $\tau > 0$ is the temperature parameter. The raw logit $z_{t,i}$ is LLM's internal score for how likely token $v_i$ at position $t$ is to appear next, given the context $w_{<t}$. Lower values of $\tau$ make sampling more deterministic, as they amplify the differences between the model's prediction score (i.e., logits), concentrating the probability mass on the most likely tokens. As a result, the model tends to consistently generate the same high-probability tokens given the same input. In contrast, higher values of $\tau$ flatten the probability distribution, increasing the likelihood of sampling less probable tokens.

The temperature parameter $\tau$ serves as a control variable for managing the trade-off between feasibility and diversity during the sampling process. Lower temperatures bias the model toward high-probability tokens, increasing feasibility, while higher temperatures promote exploration by increasing the likelihood of sampling less probable tokens, thereby promoting output diversity. Together with the attribute ordering used during training, the decoding temperature provides an additional axis for tuning the generation process to balance feasibility and diversity.

## 4.4 Evaluation Metrics

We evaluate the quality of the generated synthetic population using a set of metrics designed to assess statistical similarity, feasibility and diversity. Since the true population is not available, we follow the semi-synthetic evaluation strategy outlined in Section 3: we treat the full dataset (approximately one million records) as h-population and evaluate the model's performance in recovering its structural and distributional properties using only the h-sample (approximately 50,000 records) for training.

### 4.4.1 Distributional Similarity

To assess whether the generative model accurately captures the statistical structure of the h-population, we evaluate distributional similarity between the generated data and the h-population in terms of marginal and bivariate attribute distributions. We adopt the standardized root mean square error (SRMSE) to quantify the deviation between the empirical distribution of the h-population and that of the generated synthetic population. Let $P$ denote the empirical distribution of the h-population and $\hat{P}$ the corresponding distribution of the generated data. The SRMSE for bivariate distribution is calculated as follows:

$$\text{SRMSE}(P, \hat{P}) = \sqrt{\frac{1}{N_b} \sum_{(i,j)} \left( \frac{P(i,j) - \hat{P}(i,j)}{P(i,j)} \right)^2} \qquad (8)$$

where $N_b$ is the number of valid (non-zero) category combinations and $\hat{P}(i,j)$ is the joint probabilities of category pair $(i,j)$ in h-population and generated data, respectively. In the case of SRMSE calculated for marginal distribution, $P(i,j)$ reduces to the marginal probability $P(i)$ of categorical variable $i$. Therefore, a lower SRMSE indicates a closer alignment between distributions of the generated data and those of the h-population. However, these metrics can yield limited and potentially misleading results when applied to high-dimensional attribute combinations (Theis et al., 2015), often leading to overfitting the simplified distribution of the training data (Kim et al., 2022). Moreover, they are unable to assess the model's ability to generate sampling zeros and structural zeros— distinctive features of DGM. Therefore, in the following sections, we adopt feasibility and diversity metrics to provide a more comprehensive evaluation.

### 4.4.2 Feasibility and Diversity

We evaluate whether the generated data (i) avoids implausible combinations, referred to as structural zeros, and (ii) successfully generates plausible but previously unobserved attribute combinations, known as sampling zeros. Following the prior work (Kim and Bansal, 2023), we define feasibility and diversity as follows.

$$\text{Precision} = \frac{1}{M} \sum_{j=1}^{M} 1_{G_j \in H}, \text{Recall} = \frac{1}{N} \sum_{i=1}^{N} 1_{H_i \in G}, \text{F1 score} = \frac{2 \cdot \text{Precision} \cdot \text{Recall}}{\text{Precision} + \text{Recall}} \qquad (9)$$

where $H$ denotes the h-population with $N$ data points, and $G$ denotes the generated data with $M$ data points. The function $1(\cdot)$ is an indicator function used for counting.

Precision quantifies the feasibility of the generated data, i.e., how many of the generated attribute combinations actually exist in the h-population. A high precision score indicates that the model avoids generating structural zeros, i.e., logically infeasible or unrealistic combinations. On the other hand, recall captures the diversity of the generation data, i.e., the proportion of attribute combinations in the h-population that are covered by the generated data. Since the model is trained only on the h-sample but evaluated on the full h-population, high recall indicates that the model generalizes beyond the observed training data to recover valid but unobserved combinations (i.e., more sampling zeros). Conversely, high precision reflects the model's ability to avoid implausible combinations not found in the h-population (i.e., fewer structural zeros).

### 4.4.3 Benchmark models

To validate the performance of the proposed method, we compare six approaches:

(a) Prototypical agent: A deterministic baseline that replicates marginal distributions from the h-sample.

(b) DGM-VAE: A variational autoencoder trained to reconstruct the h-sample.

(c) DGM-WGAN: A Wasserstein GAN trained to reconstruct the h-sample.

(d) LLM-Few-shot: A proprietary LLM (GPT-4o) with few-shot learning on the h-sample (See **Appendix A**).

(e) LLM-Random: A lightweight LLM (GPT-2) fine-tuned on the h-sample using random attribute sequences.

(f) LLM-BN: A lightweight LLM (GPT-2) fine-tuned on the h-sample using attribute sequences guided by a BN.

In the prototypical agent approach, synthetic agents are generated by directly sampling from the empirical distribution of h-sample without further adjustment.

### 4.4.4 Information Sources: Semantic Relationship vs. Sample Distribution

To contextualize the evaluation of feasibility and diversity, it is critical to understand the source of information used by each benchmark model. All the proposed and benchmark models are trained (or fine-tuned) on the h-sample and evaluated against the full h-population. However, LLMs additionally incorporate prior knowledge of semantic relationships from a large-scale pretraining, which indirectly represents the population characteristics beyond those observed in the h-sample. **Table 2** summarizes this distinction.

**Table 2** Data sources leveraged during training by each approach

| Training data | DGN-VAE, WGAN | LLM-Few-shot | LLM-Random, LLM-BN |
|---|---|---|---|
| Large-scale text corpora | No | Partially (via pretraining) | Partially (via pretraining) |
| h-population | No | No | No |
| h-sample | Yes (via training) | Partially (via few-shot learning) | Yes (via fine-tuning) |

Traditional DGMs (e.g., VAEs, GANs) are trained exclusively on the h-sample and rely solely on the sample distributions. They do not incorporate any prior knowledge into the semantic relationship between attributes in the synthetic population. In contrast, proprietary LLMs (e.g., GPT-4o) with a few-shot learning (Liu et al., 2024a, 2024b) leverages semantic relationships learned from pretraining on large-scale text corpora as well as a small number of h-sample records as few-shot examples. While this enables the model to generate plausible combinations beyond the h-sample, the outputs often converge toward the most probable attribute combinations, which are most statistically or semantically dominant in the training context (i.e., both pretraining and few-shot examples), thereby limiting generation diversity. In other words, the model tends to replicate typical patterns it has frequently encountered during pretraining or in the prompt examples, while failing to explore rare but valid combinations. Our method fine-tunes a pretrained lightweight LLM using the h-sample. As a result, the model integrates both information sources, the sample distributions from the h-sample and the semantic relationships learned from the large-scale text corpora, enabling the generation of synthetic populations that are both feasible and diverse. Since the LLM is pretrained on open-domain text corpora, it may generate values that are not included in the predefined attribute categories (e.g., an unlisted occupation). However, these invalid outputs can be easily filtered using the predefined attribute set. We apply this filtering process and repeat sampling until the desired target sample size is reached (Borisov et al., 2023).

## 5. Experimental Results
### 5.1 Topological Ordering of Attributes Learned via Bayesian Network

To guide the fine-tuning of the LLM, we first learn a DAG from the training data using score-based Bayesian Network structure learning, as described in Section 4.2. The resulting structure, shown in **Figure 5**, captures the conditional dependencies among socio-demographic and travel-related attributes and defines a topological ordering that is used to structure the input sequences during training.

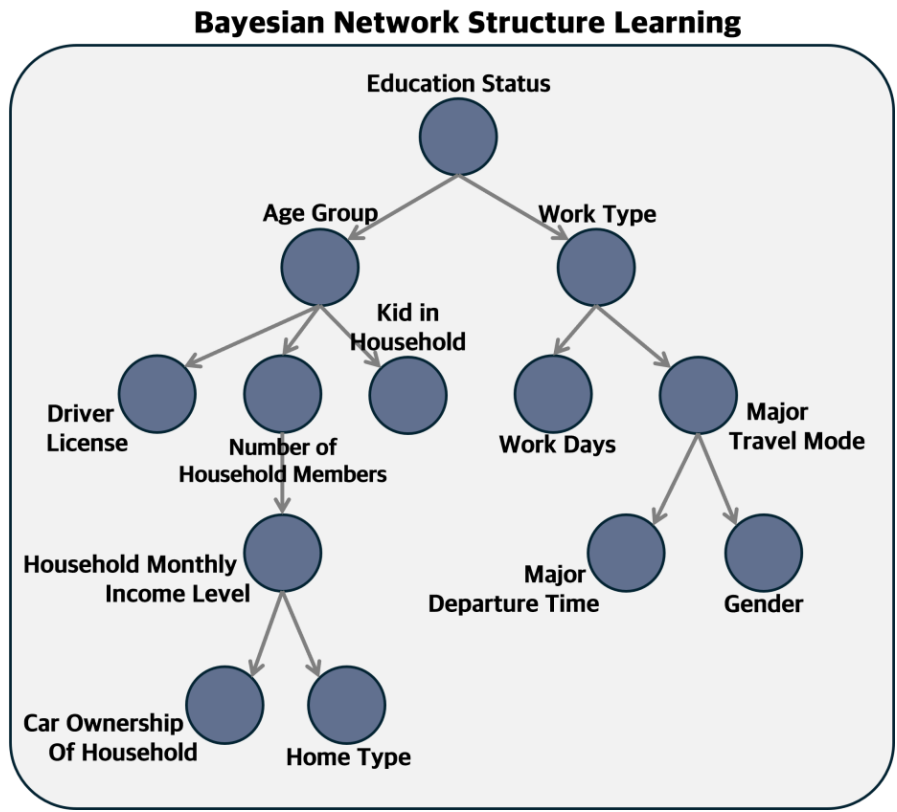

**Figure 5.** Learned DAG representing conditional dependencies among socio-demographic attributes

The learned DAG reveals a set of interpretable and plausible relationships. Education Status is positioned at the top of the hierarchy, serving as a root node that directly influences both Age Group and Work Type. Age Group plays a central role in shaping downstream socio-demographic factors, including Driver License possession, the number of Household Members, and whether children are present in the household (Kid in Household). The number of Household Members, in turn, affects Household Monthly Income Level, which subsequently determines both Household Car Ownership and Home Type. The DAG also reveals that Work Type influences the number of Work Days per week as well as Major Travel Mode. Additionally, Major Departure Time is conditionally dependent on Major Travel Mode and also exhibits a dependency on Gender.

The autoregressive generation process is constrained to follow attribute hierarchies learned from data, thereby improving the internal consistency and feasibility of generated synthetic individuals. Importantly, the learned structure is not manually imposed but extracted directly from the data, allowing our model to flexibly adapt to different population contexts without the need for handcrafted rules.

## 5.2. Evaluation Results

This section presents empirical results evaluating our proposed method for population synthesis. The central focus is not only on performance comparison, but on understanding how the generation process of LLMs can be explicitly controlled to balance feasibility (measured by precision) and diversity (measured by recall). Since feasibility and diversity exhibit a trade-off relationship, we calibrate the hyperparameters based on overall quality, F1 score. Also, the diversity of the DGMs varies according to the number of generated data. To ensure a fair comparison and reflect typical usage in population synthesis applications, each model is configured to generate data matching the size of the h-population (about one million individuals).

Regarding diversity, in addition to recall, we also evaluate the number of unique attribute combinations generated by each model. While recall accounts for how frequently each attribute combination appear in the population (i.e., giving greater weight to major combinations), the number of attribute combinations simply counts how many distinct combinations are generated, regardless of their prevalence. Therefore, even when two models achieve similar recall, differences in the number of generated combinations may indicate distinct generation behaviors. A smaller number of combinations suggests that the model concentrates on recovering major combinations, while a larger number may reflect the generation of rare or minor combinations.[1]

**Table 3** summarizes four key metrics for all benchmarked models: distributional similarity, diversity, feasibility, and the F1 score (overall quality). The evaluation results provide several noteworthy findings.

---

[1] Suppose the h-population consists of 100 individuals distributed across nine attribute combinations $\{AC_1, AC_2,…, AC_{10}\}$ with frequencies {30, 20, 10, 10, 10, 5, 5, 5, 5}. Two models generate 100 synthetic samples. Model A generates 90 feasible samples distributed as {40, 30, 10, 10, 0, 0, 0, 0, 0} covering four combinations ($AC_1$, $AC_2$, $AC_3$, and $AC_4$), with remaining 10 infeasible samples. Model B generates 90 feasible samples distributed as {0, 20, 20, 10, 10, 10, 10, 5, 5}, covering eight combinations, and produce 10 infeasible samples. Both models achieve 70% recall. This illustrates that even when models achieve the same recall values, the number of unique attribute combinations recovered can differ.

**Table 3.** Benchmarking population synthesis models: distributional similarity, diversity, and feasibility

| Model | Distributional Similarity | | Diversity | | Feasibility (Precision) | Overall Quality (F1 Score) |
|---|---|---|---|---|---|---|
| | Marg. SMRSE | Bivar. SMRSE | # of combinations | Recall | | |
| Prototypical agent | 0.008 | 0.020 | 30,837 | 56.4% | 100.0% | 72.1% |
| DGM-VAE | 0.037 | 0.089 | 331,747 | **81.5%** | 73.6% | 77.3% |
| DGM-WGAN | 0.032 | 0.094 | 263,925 | 80.8% | 81.4% | 81.1% |
| LLM-Few-shot | 0.220 | 0.617 | 40,666 | 50.2% | 84.6% | 63.0% |
| LLM-Random | 0.165 | 0.394 | 180,504 | **80.3%** | **90.8%** | **85.2%** |
| LLM-BN | 0.249 | 0.604 | 120,541 | 76.0% | **95.3%** | **84.6%** |

*Note*: Bold indicates the top two performing models.

First, the prototypical agent achieves 100% feasibility and the lowest SMRSE, but ranks second lowest in diversity, as it primarily replicates distributions observed in the h-sample. This serves as a useful baseline that reflects the information contained in the 5% h-sample without involving any generative process.

Second, despite careful prompt engineering of LLM-Few-shot using GPT-4o (See **Appendix A**), the model generated a very limited number of unique combinations, only 40,666 combinations with a recall of 50.2%. This result indicates that, without fine-tuning, LLMs struggle to capture the diversity of the population distribution when relying solely on semantic relationships acquired during pretraining. As a result, they fail to generalize well to rare but plausible combinations.

Third, although LLM-based approaches exhibit higher SRMSE than traditional DGMs (see pictorial comparison of the h-population, LLM-BN, and DGM-WGAN in **Figure 6**), LLM-Few-shot achieves slightly higher feasibility (84.6% vs. 73.6–81.4%) and LLM-Random (90.8%) and LLM-BN (95.3%) show substantially higher feasibility. High SRMSE does not translate into poor feasibility for fine-tuned LLMs, as they prioritize semantic coherence across all attributes than strict replication of marginal or bivariate distributions.

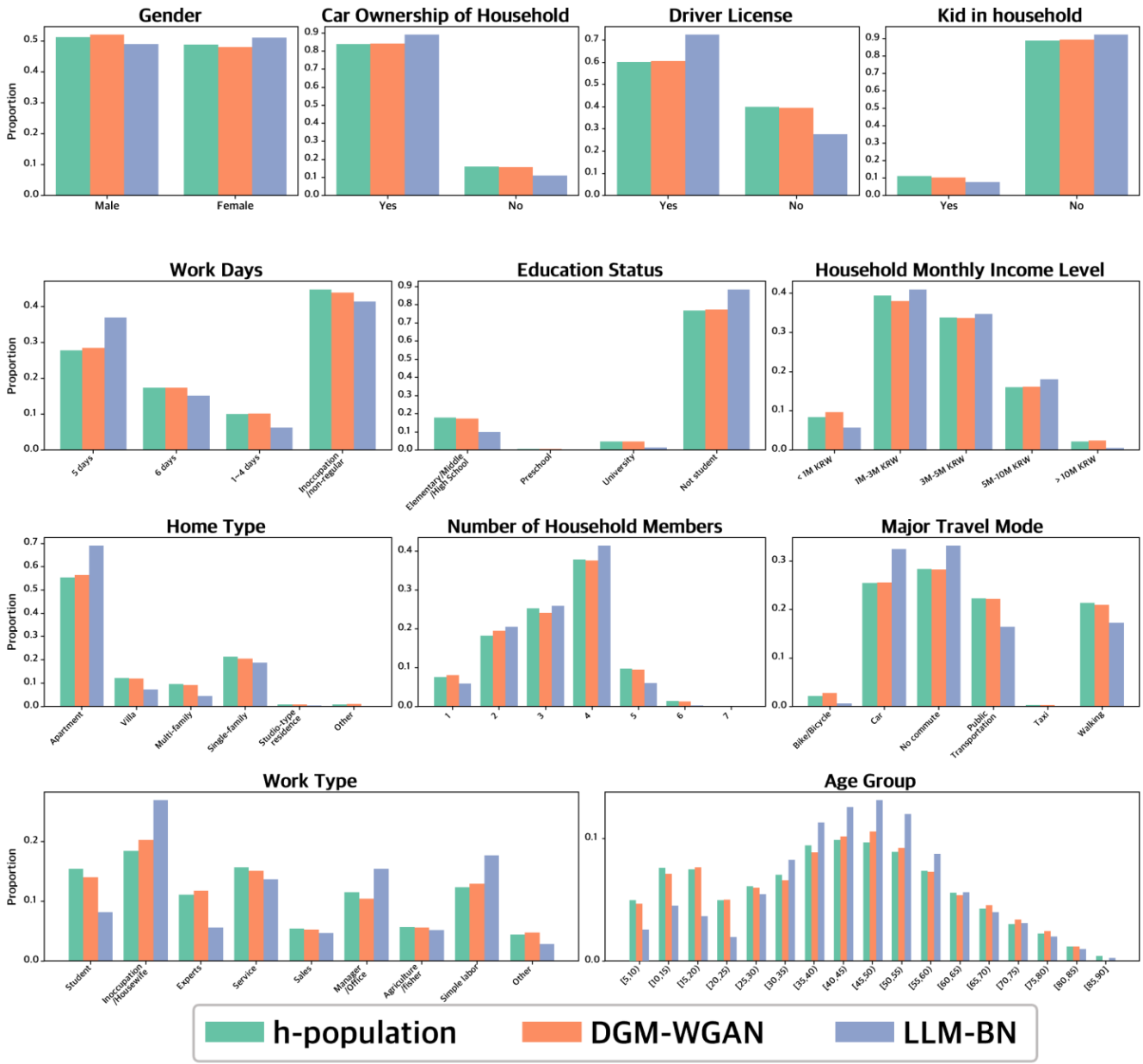

**Figure 6**. Distributional similarity between the h-population and generated samples of LLM-BN and DGM-WGAN

Fourth, while DGMs such as DGM-VAE and DGM-WGAN achieve higher recall scores (81.5% for DGM-VAE and 80.8% for DGM-WGAN) and generate a much larger number of unique combinations (331,747 and 263,925, respectively), this diversity comes at the cost of feasibility. In contrast, the proposed fine-tuned LLMs, such as LLM-Random and LLM-BN, achieve comparable recall to DGMs but with substantially higher feasibility (above 90%). As a result, the LLM-based models yield higher overall quality, as reflected in their F1 scores (85.2% and 84.6%) compared to those of the DGMs (77.3% and 81.1%). These results demonstrate that LLM-based models with fine-tuning can significantly outperform both DGMs and few-shot learning LLMs in generating feasible and diverse synthetic populations.

Fifth, although DGMs and LLM-based models exhibit similar recall values, DGMs generate nearly twice as many unique attribute combinations. This suggests that LLM-based models more successfully capture the major combinations present in the population. Such accurate identification of major combinations is informed by both the semantic relationships learned during pretraining and the sample distribution learned during fine-tuning. This may also be partially related to the higher feasibility observed in the LLM-based models.

Sixth, LLM-BN has slightly lower diversity (76.0% vs 80.3%) but higher feasibility (95.3% vs 90.8%) than LLM-Random. This is because arbitrary conditioning in LLM-Random exposes the model to diverse autoregressive paths, allowing it to generalize beyond the training data and effectively recover sampling zeros. On the other hand, LLM-BN generates a more feasible and slightly less diverse population because it focuses on generating plausible combinations consistent with the conditional dependencies (i.e., learned DAG) found in the training data. These results show that our LLM-based generation provides a controllable framework, where depending on the application, practitioners can prioritize either feasibility or diversity by adjusting the training structure or generation parameters.

Finally, beyond performance, our approach offers important practical advantages. The proposed LLM-based models are lightweight and open-source, enabling fine-tuning and inference on standard personal computing environments. This stands in contrast to proprietary LLMs (e.g., GPT-4o) or large-scale LLMs (e.g., LLaMA-3.3), which often require substantial computational resources and/or incur API usage costs. For example, our benchmark model, LLM-Few-shot, incurs approximately 15,000 USD in API costs when applied to generating synthetic populations for Seoul, which has about 9.3 million residents—an amount likely infeasible for most public agencies (see Section 5.5 for more details). Thus, the proposed method offers a viable, cost-effective, and reproducible solution for generating feasible and diverse synthetic populations in ABM applications.

## 5.3 Sensitivity Analysis for Hyperparameters

To gain deeper insight into the role of fine-tuning depth (i.e., number of training epochs) and decoding temperature in controlling the trade-off between feasibility and diversity, we conduct a sensitivity analysis. Decoding temperature regulates the randomness of attribute selection during generation, while the number of training epochs determines how closely the model aligns with the training data through fine-tuning. We use the LLM-BN model in this analysis as a representative case to illustrate how these parameters influence the balance between precision and recall, as shown in **Figure 7**.

At lower temperatures, the model prefers high-probability tokens, increasing the tendency to generate more probable individuals. As a result, the model tends to reproduce frequently observed attribute combinations in the population, leading to higher precision but lower recall, as shown in **Figure 7b**. In other words, the model favors generating "typical" attribute combinations that are well-represented in the population. In contrast, at higher temperatures, the likelihood of sampling low-probability tokens increases. This broadens the range of attribute combinations generated and leads to more sampling zeros—valid but previously unobserved combinations—thus improving recall at the expense of precision.

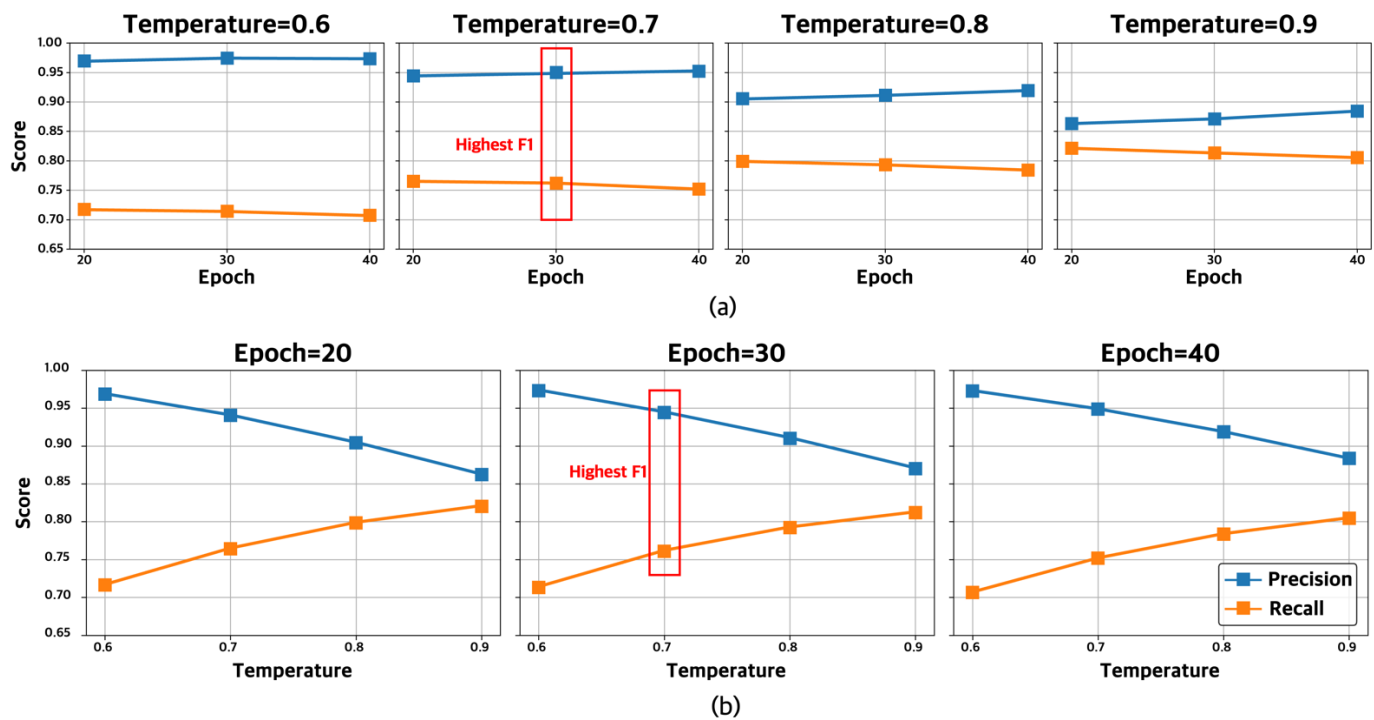

**Figure 7** Sensitivity analysis of the LLM-BN models across epochs and decoding temperatures.

    The depth of fine-tuning, defined by the number of training epochs, controls how strongly the model align with the sample distribution. As the number of epochs increases, the model adjusts its pretrained parameters more closely to replicate the attribute combinations observed in the training data. This leads to improved precision, as the model becomes increasingly confident in generating attribute combinations that are frequently represented in the sample. However, with deeper fine-tuning, the model may overfit to the sample distribution, limiting its ability to generate attribute combinations that are valid but unobserved in the training data. As a result, recall decreases with increasing epochs, as shown by the orange line in **Figure 7a**. In other words, deeper fine-tuning improves feasibility, as indicated by blue line, but comes at the cost of the reduced diversity.

    LLM-BN achieves its highest F1 score when feasibility (precision) and diversity (recall) are well balanced, as shown by the red box in **Figure 7**. Depending on the objective, users may choose this configuration or adjust parameters to emphasize either precision or recall. This interaction between the two hyperparameters indicates that one can balance the feasibility-diversity trade-off with using LLM by jointly adjusting decoding temperature and fine-tuning depth (i.e., the number of training epochs).

## 5.4 Impact of LLM model size on performance

The proposed models are developed based on a lightweight LLM, distilled GPT-2 (Sanh et al., 2019) with approximately 82 million parameters. Larger-scale LLMs may capture different levels of semantic relationship during pretraining, which could influence population synthesis performance. We compare the distilled GPT model with GPT-2 Medium and GPT-2 Large, which have larger parameter counts of 355 million and 774 million, respectively (Radford et al., 2019). The same temperature and epochs are applied to the models with different sizes of LLM. As shown in **Table 4**, replacing distilled GPT-2 with GPT-2 Medium or GPT-2 Large does not lead to improvements in model performance: while it results in a slight increase in recall, it also causes a marginal decrease in precision. This suggests that the additional semantic relationships potentially captured by the larger GPT-2 model does not contribute to the performance of the synthesized population. In other words, population synthesis may not be a highly complex task from the perspective of LLMs, and meaningful semantic relationships can be captured even with a relatively small number of parameters.

**Table 4.** Impact of LLM Size on performance

| Model | Number of parameters | Diversity | | Feasibility (Precision) | Overall Quality (F1 Score) |
| --- | --- | --- | --- | --- | --- |
| | | Number of unique combinations | Recall | | |
| Distilled GPT-2 | 82 million | 128,950 | 76.7% | 94.5% | 84.7% |
| GPT-2 Medium | 355 million | 113,240 | 75.0% | 95.5% | 84.0% |
| GPT-2 Large | 774 million | 129,970 | 76.2% | 94.4% | 84.3% |

## 5.5 Computational and API Costs

To leverage LLMs, open-source models (e.g., LLaMA-3.3) typically require expensive high-performance GPUs for both inference and fine-tuning, while proprietary models (e.g., ChatGPT-4o) incur substantial API costs for the same processes. This underscores the practicality of our proposed approach, which is based on a lightweight LLM (distilled GPT-2) that enables both fine-tuning and inference on standard personal computing environments.

To provide more detailed comparison with other LLM alternatives, we evaluate the computational times and API costs associated with fine-tuning on the h-sample (50 thousand records) and inference for the h-population (i.e., generating one million records). For local models, we assess the computational costs of fine-tuning and inference using Distilled GPT-2, GPT-2 Medium, and GPT2 Large models using a LLM-BN framework on a system equipped with a GeForce RTX 4090 GPU. For proprietary models, API costs for fine-tuning and inference are estimated based on OpenAI's pricing policy.

**Table 5** summarizes the computational time required for each step. For the Distilled GPT-2 model, LLM-BN requires 2.11 hours to complete 40 epochs of training, and 2.67 hours to generate the h-population . In contrast, the GPT-2 Medium and GPT-2 Large models require approximately 7.31 hours and 15.20 hours, respectively, to complete 40 epochs of training. Their inference times are similarly high, at over 9 and 19 hours, respectively. These results indicate that as the size of the LLM increases, the computational time required for both fine-tuning and inference grows substantially—suggesting a near-exponential increase in resource demands with LLM scale.

**Table 5.** Fine-tuning and inference time for local LLM models: Distilled GPT-2, GPT-2 Medium, and GPT-2 Large.

| Model | Method | Fine-tuning time (40 Epochs) | Inference time (One million individuals) |
|---|---|---|---|
| Distilled GPT-2 | LLM-BN | 2.12 hours | 3.84 hours |
| GPT-2 Medium | LLM-BN | 7.31 hours | 9.12 hours |
| GPT-2 Large | LLM-BN | 15.20 hours | 19.02 hours |

**Table 6** presents the estimated API costs for fine-tuning on the h-sample and performing inference for one million individuals (i.e., the h-population) and 9.3 million individuals (i.e., application to Seoul) using OpenAI's commercial API. The cost estimates are based on OpenAI's official pricing as of April 2025, with separate calculations for the fine-tuning and inference stages. The evaluated models include gpt-4.1-2025-04-14, gpt-4.1-mini-2025-04-14, gpt-4o-2024-08-06, gpt-4o-mini-2024-07-18, and gpt-3.5-turbo. The LLM-Few-shot that is benchmark model in Table 3 use the "gpt-4o-2024-08-06" without fine-tuning.

All costs are calculated based on the number of processed tokens, applying the corresponding cost per one million (1M) tokens for each model. The training data used for fine-tuning (i.e., h-sample) consists of 53,315 individuals. On average, each individual is encoded into 91 tokens, resulting in a total of 4.85 million tokens for training one epoch. For inference, we generate 1.06 million individuals (i.e., h-population). The prompt input to the API for generating each profile (i.e., "Education status is ", which is the root node in **Figure 5**) consists of 3 tokens, and the output individuals generated are encoded into 91 tokens. Accordingly, the total number of tokens for inference is calculated as 97 million (= (3+91) tokens/sample×1.06 million samples).

To further illustrate the required cost for larger-scale real-world applications, we estimate the inference cost needed to generate synthetic data for the actual population of Seoul (9.3 million individuals as of March 2025).

**Table 6.** Estimated OpenAI API costs for fine-tuning and large-scale inference

| Model | Method | Fine-tuning | | Inference | | | Total Cost | Inference for Seoul (9.3 million) |
| --- | --- | --- | --- | --- | --- | --- | --- | --- |
| | | Train (per 1M tokens) | 40 Epochs for h-sample | Input (per 1M tokens) | Output (per 1M tokens) | h-population (one million) | | |
| gpt-4.1-2025-04-14 | LLM-BN | $25.00 | $5,011.70 | $3.00 | $12.00 | $1,174.02 | $6,185.72 | $10,278.67 |
| gpt-4o-2024-08-06 | LLM-BN | $25.00 | $5,011.70 | $3.75 | $15.00 | $1,467.52 | $6,479.22 | $12,848.29 |
| gpt-3.5-turbo | LLM-BN | $8.00 | $1,603.75 | $3.00 | $6.00 | $591.81 | $2,195.56 | $5,181.36 |
| gpt-4o-2024-08-06 | LLM-Few-shot | - | - | $2.50 | $10.00 | $1,074.65 | $1,074.65 | $9,408.67 |

*Note*: The inference cost differs between fine-tuned models and non-fine-tuned models.

The total cost combining fine-tuning and inference for generating one million individuals (i.e., the h-population) is at least $2,196 for gpt-3.5-turbo, a lightweight legacy model from OpenAI, while the latest gpt-4o incurs the highest cost at approximately $6,479. Even the LLM-Few-Shot model, which uses inference only without fine-tuning, requires $1,075 for generating the h-population alone. The estimated costs become even more infeasible when generating data for the actual population size of Seoul (9.3 million individuals). In this case, gpt-3.5-turbo would cost around $5,181, while gpt-4o would incur approximately $12,848. LLM-Few-Shot would also require about $9,408 for generating the Seoul population. These results demonstrate that both fine-tuning and inference using proprietary LLMs require substantial API costs, making them likely infeasible for most public agencies.

## 6. Conclusions and Future Work

This study proposes a semantically informed population generator that fine-tunes large language models (LLMs) to generate feasible and diverse synthetic populations for activity-based models (ABMs). By incorporating topological orderings or conditional dependencies derived from a Bayesian Network (BN) along with semantic relationships during fine tuning, the proposed method improves feasibility of the synthetic population while preserving diversity.

Empirical evaluations show that the proposed method significantly outperforms traditional deep generative models (DGMs), such as variational autoencoders and generative adversarial networks, as well as proprietary LLMs (e.g., ChatGPT-4o) with few-shot learning, in terms of three key quality metrics: feasibility (precision), diversity (recall), and overall quality (F1 score). Specifically, the proposed approach achieves approximately 95% feasibility—significantly higher than the ~80% observed in DGMs—while maintaining comparable diversity, making it well-suited for practical applications. By comparing various configurations of LLM-based models, we demonstrate that the trade-off between feasibility and diversity in population synthesis can be effectively managed through design choices made during both training and generation — specifically, attribute ordering, fine-tuning depth, and decoding temperature.

Notably, the improvements in generating feasible and diverse synthetic population are achieved using a lightweight open-source LLM (Distilled GPT-2), making the method both computationally efficient and cost-effective. This makes our approach practically feasible on standard personal computing environments and adaptable to large-scale population synthesis scenarios in megacities—for example, Seoul, with a population of approximately 10 million. Therefore, the proposed method offers a scalable solution for generating representative synthetic populations without relying on expensive proprietary infrastructure—an essential consideration for transportation planners in government and public agencies.

The proposed method lays a foundation for future research and opens several promising directions. First, integrating additional sources of prior knowledge—such as domain-specific constraints (Lee et al., 2025) and human-in-the-loop feedback (Ouyang et al., 2022)—may further improve the quality and realism of generated populations. Second, extending our approach to generate daily activity schedules for synthetic populations is a promising direction to develop end-of-end LLM-based activity-based models. However, activity schedules involve much more complex attribute combinations in spatial and temporal dimensions, where semantic relationships play an even more critical role. Consequently, this extension may require larger-scale LLMs to capture higher-order semantic relationships or more advanced modeling architectures that leverage the reasoning capabilities of LLMs (Hao et al., 2024).

**CRediT authorship contribution statement**

**Sung Yoo Lim**: Conceptualization, Methodology, Data curation, Software, Formal analysis, Investigation, Writing – review & editing. **Hyunsoo Yun**: Conceptualization, Methodology, Data curation, Software, Writing – original draft, Writing – review & editing. **Prateek Bansal**: Validation, Investigation, Writing – review & editing. **Dong-Kyu Kim**: Project administration, Writing – review & editing. **Eui-Jin Kim**: Conceptualization, Methodology, Project administration, Supervision, Writing – original draft, Writing – review & editing.

**Declaration of competing interest**

The authors declare that they have no known competing financial interests or personal relationships that could have appeared to influence the work reported in this paper.

**Acknowledgements**

This work was supported by the National Research Foundation of Korea (NRF) grant funded by the Korea government (MSIT) (No.RS-2025-00520515). The authors used OpenAI's ChatGPT to correct the typos and the grammar of this manuscript. The authors verified the accuracy, validity, and appropriateness of any content generated by the language model.

**Data availability**

All code and data associated with this work are available at: https://github.com/HyunsooYun/llm-pop-synth.

**Reference**

Badu-Marfo, G., Farooq, B., Patterson, Z., 2022. Composite Travel Generative Adversarial Networks for Tabular and Sequential Population Synthesis. IEEE Trans. Intell. Transp. Syst. 23, 17976–17985. https://doi.org/10.1109/TITS.2022.3168232

Bigi, F., Rashidi, T.H., Viti, F., 2024. Synthetic Population: A Reliable Framework for Analysis for Agent-Based Modeling in Mobility. Transp. Res. Rec. 2678, 1–15. https://doi.org/10.1177/03611981241239656

Borisov, V., Seßler, K., Leemann, T., Pawelczyk, M., Kasneci, G., 2023. Language Models Are Realistic Tabular Data Generators. 11th Int. Conf. Learn. Represent. ICLR 2023 1–18.

Borysov, S.S., Rich, J., Pereira, F.C., 2019. How to generate micro-agents? A deep generative modeling approach to population synthesis. Transp. Res. Part C Emerg. Technol. 106, 73–97. https://doi.org/10.1016/j.trc.2019.07.006

Castiglione, J., Bradely, M., Gliebe, J., 2015. Activity-Based Travel Demand Models: A Primer. Transportation Research Board.

Choupani, A.A., Mamdoohi, A.R., 2016. Population Synthesis Using Iterative Proportional Fitting (IPF): A Review and Future Research. Transp. Res. Procedia 17, 223–233. https://doi.org/10.1016/j.trpro.2016.11.078

Cooper, G.F., Herskovits, E., 1992. A Bayesian Method for the Induction of Probabilistic Networks from Data. Mach. Learn. 9, 309–347. https://doi.org/10.1023/A:1022649401552

Farooq, B., Bierlaire, M., Hurtubia, R., Flötteröd, G., 2013. Simulation based population synthesis. Transp. Res. Part B Methodol. 58, 243–263. https://doi.org/10.1016/j.trb.2013.09.012

Franceschelli, G., Musolesi, M., 2024. On the Creativity of Large Language Models. AI Soc. 1–11. https://doi.org/10.1007/s00146-024-02127-3

Garrido, S., Borysov, S.S., Pereira, F.C., Rich, J., 2020. Prediction of rare feature combinations in population synthesis: Application of deep generative modelling. Transp. Res. Part C Emerg. Technol. 120, 102787. https://doi.org/10.1016/j.trc.2020.102787

Guo, J.Y., Bhat, C.R., 2007. Population synthesis for microsimulating travel behavior. Transp. Res. Rec. 92–101. https://doi.org/10.3141/2014-12

Hao, S., Gu, Y., Luo, H., Liu, T., Shao, X., Wang, X., Xie, S., Ma, H., Samavedhi, A., Gao, Q., Wang, Z., Hu, Z., 2024. LLM Reasoners: New Evaluation, Library, and Analysis of Step-by-Step Reasoning with Large Language Models. arXiv Prepr. arXiv:2404, 1–24.

Huang, L., Yu, W., Ma, W., Zhong, W., Feng, Z., Wang, H., Chen, Q., Peng, W., Feng, X., Qin, B., Liu, T., 2025. A Survey on Hallucination in Large Language Models: Principles, Taxonomy, Challenges, and Open Questions. ACM Trans. Inf. Syst. 43, 1–55. https://doi.org/10.1145/3703155

Jutras-Dubé, P., Al-Khasawneh, M.B., Yang, Z., Bas, J., Bastin, F., Cirillo, C., 2024. Copula-based transferable models for synthetic population generation. Transp. Res. Part C Emerg. Technol. 169, 104830. https://doi.org/10.1016/j.trc.2024.104830

Kim, E.J., Bansal, P., 2023. A deep generative model for feasible and diverse population synthesis. Transp. Res. Part C Emerg. Technol. 148, 104053. https://doi.org/10.1016/j.trc.2023.104053

Kim, E.J., Kim, D.K., Sohn, K., 2022. Imputing qualitative attributes for trip chains extracted from smart card data using a conditional generative adversarial network. Transp. Res. Part C Emerg. Technol. 137, 103616. https://doi.org/10.1016/j.trc.2022.103616

Kotelnikov, A., Baranchuk, D., Rubachev, I., Babenko, A., 2023. TabDDPM: Modelling Tabular Data with Diffusion Models. Proc. Mach. Learn. Res. 202, 17564–17579.

La, D.M., Vu, H.L., Kamruzzaman, L., Miller, E., 2025. Population synthesis: a problem-based review. Transp. Rev. 1–24. https://doi.org/10.1080/01441647.2025.2469069

Lederrey, G., Hillel, T., Bierlaire, M., 2022. DATGAN: Integrating expert knowledge into deep learning for synthetic tabular data. arXiv Prepr. arXiv:2203.

Lee, H., Bansal, P., Vo, K.D., Kim, E., 2025. Collaborative generative adversarial networks for fusing household travel survey and smart card data to generate heterogeneous activity schedules in urban digital twins. Transp. Res. Part C Emerg. Technol. (In Press).

Liu, T., Li, M., Yin, Y., 2024a. Can Large Language Models Capture Human Travel Behavior ? Evidence and Insights on Mode Choice. SSRN Electron. J. https://doi.org/Liu, Tianming and Li, Manzi and Yin, Yafeng, Can Large Language Models Capture Human Travel Behavior? Evidence and Insights on Mode Choice (August 26, 2024). Available at SSRN: https://ssrn.com/abstract=4937575 or http://dx.doi.org/10.2139/ssrn.4937575

Liu, T., Yang, J., Yin, Y., 2024b. Toward LLM-Agent-Based Modeling of Transportation Systems: A Conceptual Framework. arXiv Prepr. arXiv:2412.

Ouyang, L., Wu, J., Jiang, X., Almeida, D., Wainwright, C.L., Mishkin, P., Zhang, C., Agarwal, S., Slama, K., Ray, A., Schulman, J., Hilton, J., Kelton, F., Miller, L., Simens, M., Askell, A., Welinder, P., Christiano, P., Leike, J., Lowe, R., 2022. Training language models to follow instructions with human feedback. Adv. Neural Inf. Process. Syst. 35.

Pereira, R.H.M., Schwanen, T., Banister, D., 2017. Distributive justice and equity in transportation. Transp. Rev. 37, 170–191. https://doi.org/10.1080/01441647.2016.1257660

Qin, Z., Zhang, P., Wang, L., Ma, Z., 2025. LingoTrip: Spatiotemporal context prompt driven large language model for individual trip prediction. J. Public Transp. 27, 100117. https://doi.org/10.1016/j.jpubtr.2025.100117

Radford, A., Wu, J., Child, R., Luan, D., Amodei, D., Sutskever, I., 2019. Language Models are Unsupervised Multitask Learners. OpenAI blog.

Ray, P.P., 2023. ChatGPT: A comprehensive review on background, applications, key challenges, bias, ethics, limitations and future scope. Internet Things Cyber-Physical Syst. 3, 121–154. https://doi.org/10.1016/j.iotcps.2023.04.003

Rezvany, N., Bierlaire, M., Hillel, T., 2023. Simulating intra-household interactions for in- and out-of-home activity scheduling. Transp. Res. Part C Emerg. Technol. 157, 104362. https://doi.org/10.1016/j.trc.2023.104362

Saadi, I., Mustafa, A., Teller, J., Farooq, B., Cools, M., 2016. Hidden Markov Model-based population synthesis. Transp. Res. Part B Methodol. 90, 1–21. https://doi.org/10.1016/j.trb.2016.04.007

Sanh, V., Debut, L., Chaumond, J., Wolf, T., 2019. DistilBERT, a distilled version of BERT: smaller, faster, cheaper and lighter. arXiv Prepr. arXiv:1910.

Sun, L., Erath, A., 2015. A Bayesian network approach for population synthesis. Transp. Res. Part C Emerg. Technol. 61, 49–62. https://doi.org/10.1016/j.trc.2015.10.010

Theis, L., Oord, A. van den, Bethge, M., 2015. A note on the evaluation of generative models. arXiv Prepr. arXiv:1511.

Tsamardinos, I., Brown, L.E., Aliferis, C.F., 2006. The max-min hill-climbing Bayesian network structure learning algorithm. Mach. Learn. 65, 31–78. https://doi.org/10.1007/s10994-006-6889-7

Vo, K.D., Kim, E.J., Bansal, P., 2025. A novel data fusion method to leverage passively-collected mobility data in generating spatially-heterogeneous synthetic population. Transp. Res. Part B Methodol. 191, 103128. https://doi.org/10.1016/j.trb.2024.103128

W. Axhausen, K., Horni, A., Nagel, K., 2016. The Multi-Agent Transport Simulation MATSim. Ubiquity Press.

Wang, J., Jiang, R., Yang, C., Wu, Z., Onizuka, M., Shibasaki, R., Xiao, C., 2024. Large Language Models as Urban Residents: An LLM Agent Framework for Personal Mobility Generation. Adv. Neural Inf. Process. Syst. 37, 124547–124574.

Wei, J., Tay, Y., Bommasani, R., Raffel, C., Zoph, B., Borgeaud, S., Yogatama, D., Bosma, M., Zhou, D., Metzler, D., Chi, E.H., Hashimoto, T., Vinyals, O., Liang, P., Dean, J., Fedus, W., 2022. Emergent Abilities of Large Language Models. arXiv Prepr. arXiv:2206.

Zhang, Y., Zhang, K., Pang, Y., Sekimoto, Y., 2024. Agentic Large Language Models for Generating Large-Scale Urban Daily Activity Patterns. Proc. - 2024 IEEE Int. Conf. Big Data, BigData 2024 6815–6822. https://doi.org/10.1109/BigData62323.2024.10825138

Zhao, Z., Kunar, A., Birke, R., Chen, L.Y., 2021. CTAB-GAN: Effective Table Data Synthesizing. Proc. Mach. Learn. Res. 157, 97–112.

Zheng, Y., Wang, S., Zhao, J., 2021. Equality of opportunity in travel behavior prediction with deep neural networks and discrete choice models. Transp. Res. Part C Emerg. Technol. 132, 103410. https://doi.org/10.1016/j.trc.2021.103410

# Appendix A. Details for Proprietary LLM with Few-shot Learning

**Figure A.1** provides an overview of the LLM-Few-Shot model, which uses ChatGPT-4o with few-shot learning and incorporates role prompting, chain-of-thought reasoning, and function calling. This prompt was developed through extensive trial and error and was found to be the most effective in maximizing both feasibility and diversity.

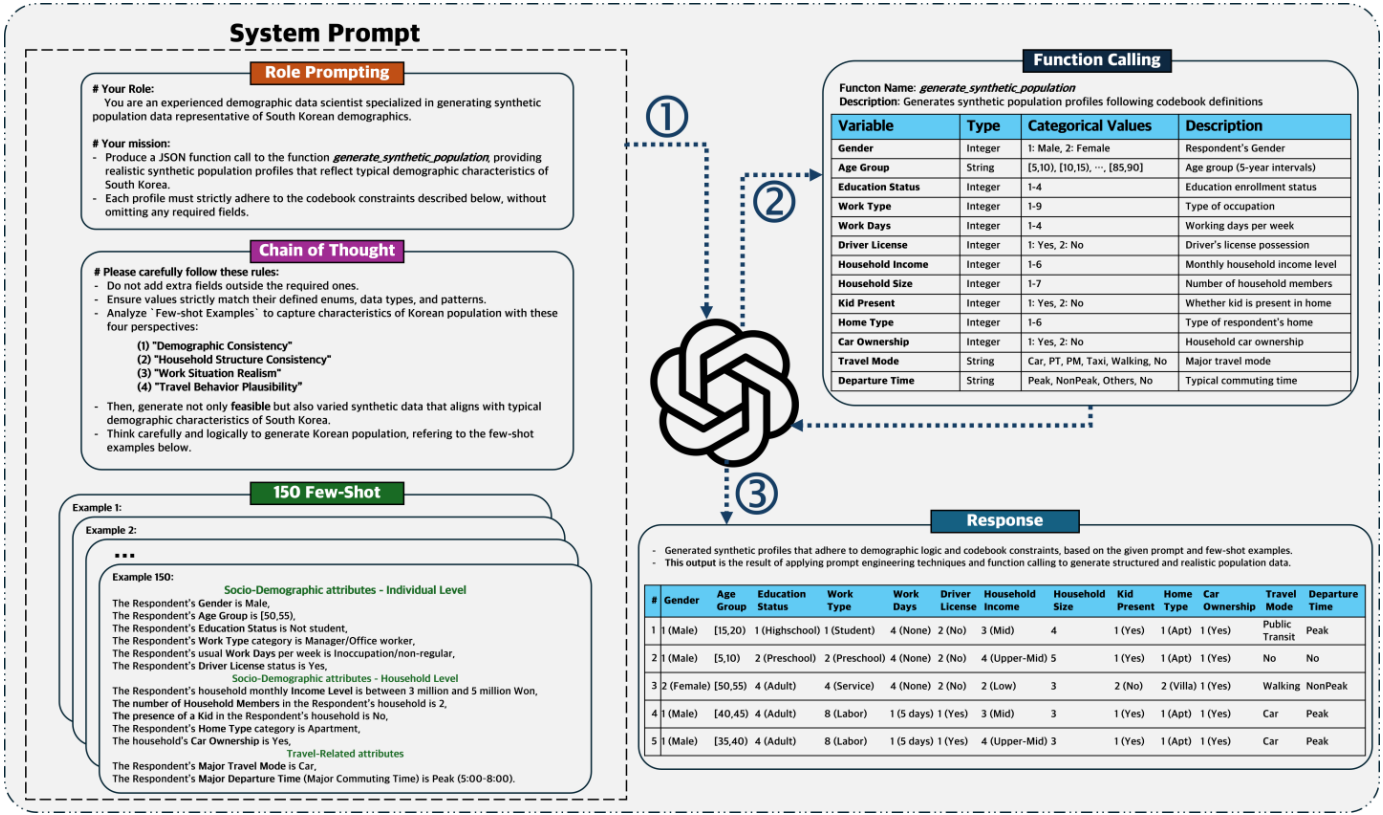

**Figure A.1**. Systematic prompt engineering framework for synthetic population generation using few-shot learning, chain-of-thought, and function calling

Role prompting is a technique that systematically guides an LLM's behavior and outputs by explicitly defining its role and setting a clear task objective. It consists of two key components. First, in Role Definition, the LLM is assigned a specific role—for example, "an experienced demographic data scientist specialized in generating synthetic data reflecting South Korean demographics." Second, in Mission Definition, the LLM is instructed to produce outputs in a structured JSON format that adheres to a predefined schema (i.e., tabular format for synthetic population data). These definitions are embedded in the system prompt and serve as the foundation for establishing task context and objectives.

      Chain-of-Thought (CoT) prompting is used to guide the LLM through a structured reasoning process. For population synthesis, we apply CoT by incorporating four explicit reasoning steps to ensure semantic consistency among attributes:

· **Demographic consistency**: Ensuring accurate representation of key socio-demographic attributes (e.g., age, gender, education enrollment status) as observed in South Korea.

- **Household structure consistency**: Maintaining logical coherence between household attributes (e.g., the number of household members, presence of children)
- **Work situation realism**: Aligning work-related attributes (e.g., work type, working days per week) to reflect plausible relationships.
- **Travel behavior plausibility**: Ensuring that travel modes and departure times logically correspond with socio-demographic profiles.

Based on these reasoning steps, the LLM analyzes the few-shot examples provided from the h-sample. Few-shot learning is a prompting technique that enables the LLM to perform specific tasks without additional fine-tuning. Providing a small number of examples improves the overall quality of the generated synthetic population (Liu et al., 2024b). For each API call, we randomly sample 150 individuals from the h-sample and present them as few-shot examples. This randomization introduces variability in each call, thereby promoting diverse synthetic population outputs. In few-shot learning, attributes are grouped based on semantic hierarchies and presented in the following order: individual attributes (*Gender, Age Group, Education Status, Work type, Work Days, Driver License*) → household attributes (*Household Income, Household Size, Kid in Household, Home Type, and Car Ownership*) → travel-related attributes (*Major Travel mode, and Major Departure Time*). By organizing attributes by subject and semantic structure, the LLM is better guided to recognize meaningful semantic relationships.

Finally, OpenAI's *Function Calling* is used to enforce output formatting, requiring the LLM to return responses in a predefined JSON schema with expected attribute values. This schema explicitly defines the variable names listed above along with their expected categorical values and detailed descriptions as shown in FigureA.1. Furthermore, it is another form of prompt technique, where function schemas act as precise instructions guiding the LLM to generate strictly structured outputs. Thereby, this structured format converts unstructured text output into standardized tabular data, facilitating downstream processing.